\documentstyle[aclap,psfig]{article}
\title{A Word-to-Word Model of Translational Equivalence}
\author{I. Dan Melamed \\ 
Dept. of Computer and Information Science \\
University of Pennsylvania \\ 
Philadelphia, PA, 19104, U.S.A. \\
{\tt melamed@unagi.cis.upenn.edu}}

\date{}

\begin{document}
\maketitle

\begin{abstract}

Many multilingual NLP applications need to translate words between
different languages, but cannot afford the computational expense of
inducing or applying a full translation model.  For these
applications, we have designed a fast algorithm for estimating a
partial translation model, which accounts for translational
equivalence only at the word level.  The model's precision/recall
trade-off can be directly controlled via one threshold parameter.
This feature makes the model more suitable for applications that are
not fully statistical.  The model's hidden parameters can be easily
conditioned on information extrinsic to the model, providing an easy
way to integrate pre-existing knowledge such as part-of-speech,
dictionaries, word order, etc..  Our model can link word tokens in
parallel texts as well as other translation models in the literature.
Unlike other translation models, it can automatically produce
dictionary-sized translation lexicons, and it can do so with over 99\%
accuracy.
\end{abstract}

\section{Introduction} 

Over the past decade, researchers at IBM have developed a series of
increasingly sophisticated statistical models for machine translation
\cite{candide,ibm90,ibm}.  However, the IBM models, which attempt to
capture a broad range of translation phenomena, are computationally
expensive to apply.  Table look-up using an explicit translation
lexicon is sufficient and preferable for many multilingual NLP
applications, including ``crummy'' MT on the World Wide Web
\cite{crummy}, certain machine-assisted translation tools
(e.g. \cite{transcheck,adomit}), concordancing for bilingual
lexicography \cite{catiz,wordcorr}, computer-assisted language
learning, corpus linguistics \cite{corpling}, and cross-lingual
information retrieval \cite{mlir}.

In this paper, we present a fast method for inducing accurate
translation lexicons.  The method assumes that words are translated
one-to-one.  This assumption reduces the explanatory power of our
model in comparison to the IBM models, but, as shown in
Section~\ref{linkalg}, it helps us to avoid what we call indirect
associations, a major source of errors in other models.
Section~\ref{linkalg} also shows how the one-to-one assumption enables us
to use a new greedy competitive linking algorithm for re-estimating the
model's parameters, instead of more expensive algorithms that
consider a much larger set of word correspondence possibilities.  The
model uses two hidden parameters to estimate the confidence of its own
predictions.  The confidence estimates enable direct control of the
balance between the model's precision and recall via a simple
threshold.  The hidden parameters can be conditioned on prior
knowledge about the bitext to improve the model's accuracy.

\section{Co-occurrence}
With the exception of \cite{funb}, previous methods for automatically
constructing statistical translation models begin by looking at word
co-occurrence frequencies in bitexts
\cite{wordcorr,kumano,funa,mel95}.  A {\bf bitext} comprises a pair of
texts in two languages, where each text is a translation of the other.
Word co-occurrence can be defined in various ways.  The most common
way is to divide each half of the bitext into an equal number of
segments and to align the segments so that each pair of segments $S_i$
and $T_i$ are translations of each other \cite{align,simr}.  Then, two
word tokens $(u,v)$ are said to {\bf co-occur} in the aligned segment
pair $i$ if $u \in S_i$ and $v \in T_i$.  The co-occurrence relation
can also be based on distance in a bitext space, which is a more
general representations of bitext correspondence
\cite{wordalign,anlp97}, or it can be restricted to words pairs that
satisfy some matching predicate, which can be extrinsic to the model
\cite{mel95,portable}.

\section{The Basic Word-to-Word Model}

Our translation model consists of the hidden parameters $\lambda^+$
and $\lambda^-$, and likelihood ratios $L({\bf u,v})$.  The two hidden
parameters are the probabilities of the model generating true and
false positives in the data.  $L({\bf u,v})$ represents the likelihood
that ${\bf u}$ and ${\bf v}$ can be mutual translations.  For each
co-occurring pair of word types ${\bf u}$ and ${\bf v}$, these
likelihoods are initially set proportional to their co-occurrence
frequency $n_{({\bf u,v})}$ and inversely proportional to their
marginal frequencies $n_{({\bf u})}$ and $n_{({\bf v})}$ \hspace*{-.1in}
\footnote{The co-occurrence frequency of a word type pair is simply
the number of times the pair co-occurs in the corpus.  However,
$n_{({\bf u})} = \sum_{\bf v} n_{({\bf u,v})}$, which is {\em not} the
same as the frequency of {\bf u}, because each token of {\bf u} can
co-occur with several different{\bf v}'s.}, following
\cite{dunn}\footnote{We could just as easily use other symmetric
``association'' measures, such as $\phi^2$ \cite{wordcorr} or the Dice
coefficient \cite{smadja}.}.  When the $L({\bf u,v})$ are
re-estimated, the model's hidden parameters come into play.

After initialization, the model induction algorithm iterates:
\begin{enumerate}
\item Find a set of ``links'' among word tokens in the bitext, using the
likelihood ratios and the competitive linking algorithm.
\item Use the links to re-estimate $\lambda^+$, $\lambda^-$, and the
likelihood ratios.
\item Repeat from Step 1 until the model converges to the desired
degree.  
\end{enumerate}
The competitive linking algorithm and its one-to-one assumption are
detailed in Section~\ref{linkalg}.  Section~\ref{paramest} explains
how to re-estimate the model parameters.

\subsection{Competitive Linking Algorithm}
\label{linkalg}

\begin{figure}[htb]
\centerline{\psfig{figure=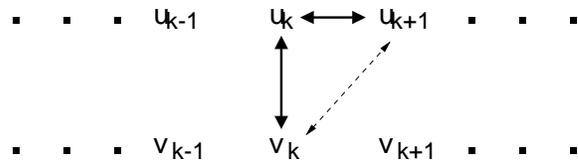,width=2.8in}}
\caption{{\em $u_{k}$ and $v_{k}$ often co-occur, as do $u_{k}$ and
$u_{k+1}$.  The direct association between $u_{k}$ and $v_{k}$, and
the direct association between $u_{k}$ and $u_{k+1}$ give rise to an
indirect association between $v_{k}$ and $u_{k+1}$.}\label{dep}}
\end{figure}

The competitive linking algorithm is designed to overcome the problem
of indirect associations, illustrated in Figure~\ref{dep}.
The sequences of $u$'s and $v$'s represent corresponding regions of a
bitext.  If ${\bf u}_k$ and ${\bf v}_k$ co-occur much more often than
expected by chance, then any reasonable model will deem them likely to
be mutual translations.  If ${\bf u}_k$ and ${\bf v}_k$ are indeed
mutual translations, then their tendency to co-occur is called a {\bf
direct association}.  Now, suppose that ${\bf u}_k$ and ${\bf
u}_{k+1}$ often co-occur within their language.  Then ${\bf v}_k$ and
${\bf u}_{k+1}$ will also co-occur more often than expected by chance.
The arrow connecting ${\bf v}_k$ and ${\bf u}_{k+1}$ in
Figure~\ref{dep} represents an {\bf indirect association}, since the
association between ${\bf v}_k$ and ${\bf u}_{k+1}$ arises only by
virtue of the association between each of them and ${\bf u}_k$.
Models of translational equivalence that are ignorant of indirect
associations have ``a tendency ... to be confused by collocates''
\cite{wordalign}.

\label{paramest}
\begin{figure*}[htb]
\centering
\begin{tabular}{|rcl|}
\hline
$n_{\bf (u,v)}$  & = & frequency of co-occurrence between word types
${\bf u}$ and ${\bf v}$ \\
$N$ & = & $\sum_{\bf (u,v)} n_{\bf (u,v)} = $ total number of
co-occurrences in the bitext \\
$k_{\bf (u,v)}$  & = & frequency of links between word types ${\bf u}$
and ${\bf v}$ \\
$K$ & = & $\sum_{\bf (u,v)} k_{\bf (u,v)} = $ total number of links in
the bitext \\
$\tau$ & = & $\Pr$( mutual translations $|$ co-occurrence ) \\
$\lambda$ & =  & $\Pr$( link $|$ co-occurrence ) \\
$\lambda^+$ & =  & $\Pr$( link $|$ co-occurrence of mutual translations ) \\
$\lambda^-$ & =  & $\Pr$( link $|$ co-occurrence of not mutual translations ) \\
$B(k | n, p)$ & = & $\Pr$( $k | n , p$), where $k$ has a binomial
distribution with parameters $n$ and $p$ \\
\hline
\multicolumn{3}{|c|}{N.B.: $\lambda^+$ and $\lambda^-$
need not sum to 1, because they are conditioned on different events.} \\
\hline
\end{tabular}
\caption{{\em Variables used to estimate the model parameters.}\label{vars}}
\end{figure*}

Fortunately, indirect associations are usually not difficult to identify,
because they tend to be weaker than the direct associations on which
they are based \cite{amta}.  The majority of indirect associations can
be filtered out by a simple competition heuristic: Whenever several
word tokens $u_i$ in one half of the bitext co-occur with a particular
word token $v$ in the other half of the bitext, the word that is most
likely to be $v$'s translation is the one for which the likelihood
$L({\bf u,v})$ of translational equivalence is highest.  The
competitive linking algorithm implements this heuristic:
\begin{enumerate}
\item Discard all likelihood scores for word types deemed unlikely to
be mutual translations, i.e. all $L({\bf u, v)} < 1$.  This step
significantly reduces the computational burden of the algorithm.  It
is analogous to the step in other translation model induction
algorithms that sets all probabilities below a certain threshold to
negligible values \cite{ibm90,wordalign,chen}.  To retain word type pairs
that are at least twice as likely to be mutual translations than not,
the threshold can be raised to~2.  Conversely, the threshold can be
lowered to buy more coverage at the cost of a larger model that will
converge more slowly.
\item Sort all remaining likelihood estimates $L({\bf u, v)}$ from highest
to lowest.
\item Find ${\bf u}$ and ${\bf v}$ such that the likelihood
ratio $L({\bf u, v})$ is highest.  Token pairs of these types would be
the winners in any competitions involving ${\bf u}$ or ${\bf v}$.
\item Link all token pairs $({\bf u, v})$ in the bitext.
\item The one-to-one assumption means that linked words cannot be
linked again.  Therefore, remove all linked word tokens from
their respective texts.
\item If there is another co-occurring word token pair $(u, v)$ such
that $L({\bf u, v})$ exists, then repeat from Step~3.
\end{enumerate}

The competitive linking algorithm is more greedy than algorithms that
try to find a {\em set} of link types that are jointly most probable
over some segment of the bitext.  In practice, our linking algorithm
can be implemented so that its worst-case running time is $O(lm)$,
where $l$ and $m$ are the lengths of the aligned segments.

The simplicity of the competitive linking algorithm depends on the
{\bf one-to-one assumption}: Each word translates to at most one other
word.  Certainly, there are cases where this assumption is false.
We prefer not to model those cases, in order to achieve higher
accuracy with less effort on the cases where the assumption is true.

\subsection{Parameter Estimation}

\begin{figure}[ht]
\centerline{\psfig{figure=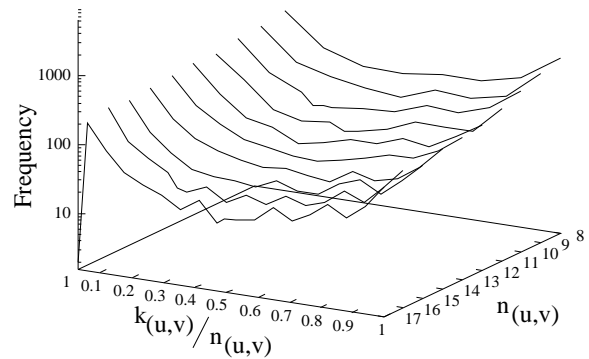,width=3in}}
\caption{{\em A fragment of the joint frequency $(k_{{\bf (u,v)}} /
n_{{\bf (u,v)}}, n_{{\bf (u,v)}})$.  Note that the frequencies are
plotted on a log scale --- the bimodality is quite sharp.}\label{bimodal}}
\end{figure}

The purpose of the competitive linking algorithm is to help us
re-estimate the model parameters.  The variables that we use in our
estimation are summarized in Figure~\ref{vars}.  The linking algorithm
produces a set of links between word tokens in the bitext.  We define
a {\bf link token} to be an ordered pair of word tokens, one from each
half of the bitext.  A {\bf link type} is an ordered pair of word
types.  Let $n_{{\bf (u,v)}}$ be the co-occurrence frequency of ${\bf
u}$ and ${\bf v}$ and $k_{({\bf u,v})}$ be the number of links between
tokens of ${\bf u}$ and ${\bf v}$\footnote{Note that $k_{({\bf u,v})}$
depends on the linking algorithm, but $n_{{\bf (u,v)}}$ is a constant
property of the bitext.}.  An important property of the competitive
linking algorithm is that the ratio $k_{{\bf (u,v)}} / n_{{\bf
(u,v)}}$ tends to be very high if ${\bf u}$ and ${\bf v}$ are mutual
translations, and quite low if they are not.  The bimodality of this
ratio for several values of $n_{{\bf (u,v)}}$ is illustrated in
Figure~\ref{bimodal}.  This figure was plotted after the model's first
iteration over
300000 aligned sentence pairs from the Canadian Hansard bitext.  Note
that the frequencies are plotted on a log scale --- the bimodality is
quite sharp.

The linking algorithm creates all the links of a given type
independently of each other, so the number $k_{{\bf (u,v)}}$ of links
connecting word types ${\bf u}$ and ${\bf v}$ has a binomial
distribution with parameters $n_{{\bf (u,v)}}$ and $p_{{\bf (u,v)}}$.
If ${\bf u}$ and ${\bf v}$ are mutual translations, then $p_{{\bf
(u,v)}}$ tends to a relatively high probability, which we will call
$\lambda^+$.  If ${\bf u}$ and ${\bf v}$ are not mutual translations,
then $p_{{\bf (u,v)}}$ tends to a very low probability, which we will
call $\lambda^-$.  $\lambda^+$ and $\lambda^-$ correspond to the two
peaks in the frequency distribution of $k_{{\bf (u,v)}} / n_{{\bf
(u,v)}}$ in Figure~2.  The two parameters can also be interpreted as
the percentage of true and false positives.  If the translation in the
bitext is consistent and the model is accurate, then $\lambda^+$
should be near 1 and $\lambda^-$ should be near 0.  

To find the most probable values of the hidden model parameters
$\lambda^+$ and $\lambda^-$, we adopt the standard method of
maximum likelihood estimation, and find the values that maximize the
probability of the link frequency distributions. The one-to-one
assumption implies independence between different link types, so that
\begin{equation}
\label{prdata}
\Pr ( links | model ) = \prod_{{\bf u,v}} \Pr(k_{({\bf u,v})} |
n_{({\bf u,v})}, \lambda^+, \lambda^-).
\end{equation}
The factors on the right-hand side of Equation~\ref{prdata} can be
written explicitly with the help of a mixture coefficient.  Let $\tau$
be the probability that an arbitrary co-occurring pair of word types
are mutual translations.  Let $B(k|n,p)$ denote the probability that
$k$ links are observed out of $n$ co-occurrences, where $k$ has a
binomial distribution with parameters $n$ and $p$.  Then the
probability that ${\bf u}$ and ${\bf v}$ are linked $k_{({\bf
u,v})}$ times out of $n_{({\bf u,v})}$ co-occurrences is a mixture of
two binomials:
\begin{eqnarray}
\label{twobin}
\lefteqn{\hspace{-.3in} \Pr(k_{({\bf u,v})} | n_{({\bf u,v})} , \lambda^+, \lambda^-) \hspace{.1in} =} \\
& = & \hspace{.3in} \tau B(k_{({\bf u,v})} | n_{({\bf u,v})}, \lambda^+) \nonumber \\
& + & (1 - \tau) B(k_{({\bf u,v})} | n_{({\bf u,v})}, \lambda^-) \hspace{.3in} . \nonumber
\end{eqnarray}

One more variable allows us to express $\tau$ in terms of
$\lambda^+$ and $\lambda^-$: Let $\lambda$ be the probability that an
arbitrary co-occuring pair of word tokens will be linked, regardless
of whether they are mutual translations.  Since $\tau$ is constant
over all word types, it also represents the probability that an
arbitrary co-occurring pair of word {\em tokens} are mutual
translations.  Therefore,
\begin{equation}
\label{L1}
\lambda = \tau \lambda^+ + (1 - \tau) \lambda^-.
\end{equation}
$\lambda$ can also be estimated empirically.  Let $K$ be the total
number of links in the bitext and let $N$ be the total number of
co-occuring word token pairs: \( K = \sum_{({\bf u,v})} k_{({\bf
u,v})}\), \( N = \sum_{({\bf u,v})} n_{({\bf u,v})}\).  By definition,
\begin{equation}
\label{L2}
\lambda = K / N.
\end{equation}
Equating the right-hand sides of Equations (\ref{L1}) and (\ref{L2})
and rearranging the terms, we get:
\begin{equation}
\label{tau}
\tau = \frac{K / N - \lambda^-}{\lambda^+ - \lambda^-}.
\end{equation}
Since $\tau$ is now a function of $\lambda^+$ and $\lambda^-$, only
the latter two variables represent degrees of freedom in the model.

\begin{figure}[ht]
\centerline{\psfig{figure=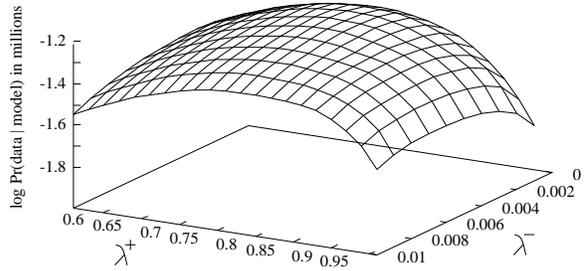,width=3in}}
\caption{{\em $\Pr(links|model)$ has only one global maximum in the
region of interest.}\label{surface}}
\end{figure}
The probability function expressed by Equations~\ref{prdata}
and~\ref{twobin} has many local maxima.  In practice, these local
maxima are like pebbles on a mountain, invisible at low resolution.
We computed Equation~\ref{prdata} over various combinations of
$\lambda^+$ and $\lambda^-$ after the model's first iteration over
300000 aligned sentence pairs from the Canadian Hansard bitext.
Figure~4 shows that the region of interest in the parameter space,
where $1 > \lambda^+ > \lambda > \lambda^- > 0$, has only one clearly
visible global maximum.  This global maximum can be found by standard
hill-climbing methods, as long as the step size is large enough to
avoid getting stuck on the pebbles.

Given estimates for $\lambda^+$ and $\lambda^-$, we can compute
$B(k_{{\bf u,v}} | n_{{\bf u,v}}, \lambda^+)$ and $B(k_{{\bf u,v}} |
n_{{\bf u,v}}, \lambda^-)$.  These are probabilities that $k_{({\bf
u,v})}$ links were generated by an algorithm that generates correct
links and by an algorithm that generates incorrect links, respectively,
out of $n_{({\bf u,v})}$ co-occurrences.  The ratio of these
probabilities is the likelihood ratio in favor of ${\bf u}$ and ${\bf
v}$ being mutual translations, for all ${\bf u}$ and ${\bf v}$:
\begin{equation}
\label{ll}
L({\bf u,v}) = \frac{B(k_{{\bf u,v}} | n_{{\bf u,v}},
\lambda^+)}{B(k_{{\bf u,v}} | n_{{\bf u,v}}, \lambda^-)}.
\end{equation}

\section{Class-Based Word-to-Word Models}

In the basic word-to-word model, the hidden parameters $\lambda^+$ and
$\lambda^-$ depend only on the distributions of link frequencies
generated by the competitive linking algorithm.  More accurate models
can be induced by taking into account various features of the linked
tokens.  For example, frequent words are translated less consistently
than rare words \cite{sement}.  To account for this difference, we can
estimate separate values of $\lambda^+$ and $\lambda^-$ for different
ranges of $n_{\bf (u,v)}$.  Similarly, the hidden parameters can be
conditioned on the linked parts of speech.  Word order can be taken
into account by conditioning the hidden parameters on the relative
positions of linked word tokens in their respective sentences.  Just
as easily, we can model links that coincide with entries in a
pre-existing translation lexicon separately from those that do not.
This method of incorporating dictionary information seems simpler than
the method proposed by Brown et al. for their models \cite{dictdata}.
When the hidden parameters are conditioned on different link classes,
the estimation method does not change; it is just repeated for each
link class.

\section{Evaluation}
\label{eval}
A word-to-word model of translational equivalence can be evaluated
either over types or over tokens.  It is impossible to replicate the
experiments used to evaluate other translation models in the
literature, because neither the models nor the programs that induce
them are generally available.  For each kind of evaluation, we have
found one case where we can come close.

We induced a two-class word-to-word model of translational equivalence
from 13 million words of the Canadian Hansards, aligned using the
method in \cite{align}.  One class represented content-word links and
the other represented function-word links\footnote{Since function
words can be identified by table look-up, no POS-tagger was
involved.}.  Link types with negative log-likelihood were discarded
after each iteration.  Both classes' parameters converged after six
iterations.  The value of class-based models was demonstrated by the
differences between the hidden parameters for the two classes.
$(\lambda^+, \lambda^-)$ converged at (.78,.00016) for content-class
links and at (.43,.000094) for function-class links.

\subsection{Link Types}
The most direct way to evaluate the link types in a word-level model
of translational equivalence is to treat each link type as a candidate
translation lexicon entry, and to measure precision and recall.  This
evaluation criterion carries much practical import, because many of
the applications mentioned in Section~1 depend on accurate
broad-coverage translation lexicons.  Machine readable bilingual
dictionaries, even when they are available, have only limited coverage
and rarely include domain-specific terms \cite{anlp97}.

We define the recall of a word-to-word translation model as the
fraction of the bitext vocabulary represented in the model.
Translation model precision is a more thorny issue, because people
disagree about the degree to which context should play a role in
judgements of translational equivalence.  We hand-evaluated the
precision of the link types in our model in the context of the bitext
from which the model was induced, using a simple bilingual
concordancer.  A link type ${\bf (u,v)}$ was considered correct if $u$
and $v$ ever co-occurred as direct translations of each other.  Where
the one-to-one assumption failed, but a link type captured part of a
correct translation, it was judged ``incomplete.''  Whether incomplete
links are correct or incorrect depends on the application.

\begin{figure}[htb]
\centerline{\psfig{figure=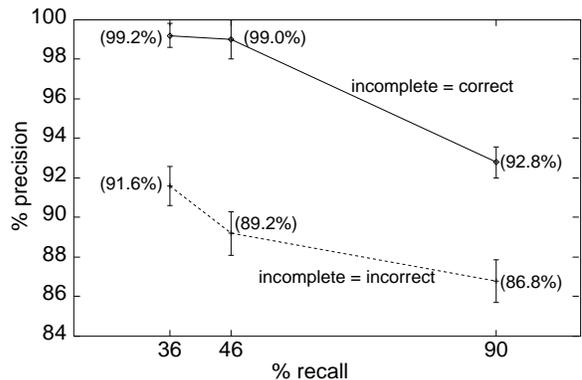,width=3in}}
\caption{{\em Link type precision with 95\% confidence intervals at
varying levels of recall.}}
\label{results}
\end{figure}
We evaluated five random samples of 100 link types each at three
levels of recall.  For our bitext, recall of 36\%, 46\% and 90\%
corresponded to translation lexicons containing 32274, 43075 and 88633
words, respectively.  Figure~\ref{results} shows the precision of the
model with 95\% confidence intervals.  The upper curve represents
precision when incomplete links are considered correct, and the lower
when they are considered incorrect.  On the former metric, our model
can generate translation lexicons with precision and recall both
exceeding 90\%, as well as dictionary-sized translation lexicons that
are over 99\% correct.  

Though some have tried, it is not clear how to extract such accurate
lexicons from other published translation models.  Part of the
difficulty stems from the implicit assumption in other models that
each word has only one sense.  Each word is assigned the same unit of
probability mass, which the model distributes over all candidate
translations.  The correct translations of a word that has several
correct translations will be assigned a lower probability than the
correct translation of a word that has only one correct translation.
This imbalance foils thresholding strategies, clever as they might be
\cite{wordcorr,wuxia,chen}.  The likelihoods in the word-to-word model
remain unnormalized, so they do not compete.

The word-to-word model maintains high precision even given much less
training data.  Resnik \& Melamed (1997) report that the model
produced translation lexicons with 94\% precision and 30\% recall,
when trained on French/English software manuals totaling about 400,000
words.  The model was also used to induce a translation lexicon from a
6200-word corpus of French/English weather reports.  Nasr (1997)
reported that the translation lexicon that our model induced from this
tiny bitext accounted for 30\% of the word types with precision
between 84\% and 90\%.  Recall drops when there is less training data,
because the model refuses to make predictions that it cannot make with
confidence.  For many applications, this is the desired behavior.

\subsection{Link Tokens}

\begin{table}[htb]
\centering
\begin{tabular}{|c|c|c|c|}
\hline
type of error & errors made by & errors made \\
		& IBM Model 2 & by our model \\
\hline
\hline
wrong link	&	32	&	 7	\\
missing link	&	12	&	36	\\
partial link	&	 7	&	10	\\
class conflict	&	--	&	 5	\\ 
tokenization	&	 3	&	 2	\\
paraphrase	& 	39	&	36	\\
\hline
TOTAL		&	93	&	96	\\ 
\hline
\end{tabular}
\caption{{\em Erroneous link tokens generated by two translation models.} \label{errors}}
\end{table}

The most detailed evaluation of link tokens to date was performed by
\cite{hannan}, who trained Brown et al.'s Model 2 on 74 million words
of the Canadian Hansards.  These authors kindly provided us with the
links generated by that model in 51 aligned sentences from a held-out
test set.  We generated links in the same 51 sentences using our
two-class word-to-word model, and manually evaluated the content-word
links from both models.  The IBM models are directional; i.e. they
posit the English words that gave rise to each French word, but ignore
the distribution of the English words.  Therefore, we ignored English
words that were linked to nothing.

The errors are classified in Table~\ref{errors}.  The ``wrong link''
and ``missing link'' error categories should be self-explanatory.
``Partial links'' are those where one French word resulted from
multiple English words, but the model only links the French word to
one of its English sources.  ``Class conflict'' errors resulted from
our model's refusal to link content words with function words.
Usually, this is the desired behavior, but words like English
auxiliary verbs are sometimes used as content words, giving rise to
content words in French.  Such errors could be overcome by a model
that classifies each word token, for example using a part-of-speech
tagger, instead of assigning the same class to all tokens of a given
type.  The bitext pre-processor for our word-to-word model split
hyphenated words, but Macklovitch \& Hannan's preprocessor did not.
In some cases, hyphenated words were easier to link correctly; in
other cases they were more difficult.  Both models made some errors
because of this tokenization problem, albeit in different places.  The
``paraphrase'' category covers all link errors that resulted from
paraphrases in the translation.  Neither IBM's Model 2 nor our model
is capable of linking multi-word sequences to multi-word sequences,
and this was the biggest source of error for both models.

The test sample contained only about 400 content words\footnote{The
exact number depends on the tokenization method.}, and the links
for both models were evaluated post-hoc by only one evaluator.
Nevertheless, it appears that our word-to-word model with only two link
classes does not perform any worse than IBM's Model 2, even though the
word-to-word model was trained on less than one fifth the amount of
data that was used to train the IBM model.  Since it doesn't store
indirect associations, our word-to-word model contained an average of
4.5 French words for every English word.  Such a compact model
requires relatively little computational effort to induce and to apply.

\begin{figure}[htb]
\centerline{\psfig{figure=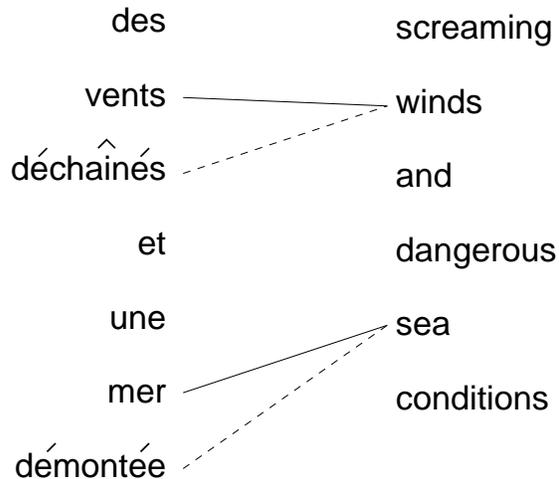,width=2.8in}}
\caption{{\em An example of the different sorts of errors made by the
word-to-word model and the IBM Model 2.  Only content-class links are
shown.  Solid lines are links generated by both models; dashed lines
are links generated by the IBM model only.  Neither model generates
the correct links (d\'{e}cha\^{i}n\'{e}s, screaming) and
(d\'{e}mont\'{e}e, dangerous). \label{links}}}
\end{figure}
In addition to the quantitative differences between the word-to-word
model and the IBM model, there is an important qualitative difference,
illustrated in Figure~\ref{links}.  As shown in Table~\ref{errors},
the most common kind of error for the word-to-word model was a missing
link, whereas the most common error for IBM's Model 2 was a wrong
link.  Missing links are more informative: they indicate where the
model has failed.  The level at which the model trusts its own
judgement can be varied directly by changing the likelihood cutoff in
Step 1 of the competitive linking algorithm.  Each application of the
word-to-word model can choose its own balance between link token
precision and recall.  An application that calls on the word-to-word
model to link words in a bitext could treat unlinked words differently
from linked words, and avoid basing subsequent decisions on uncertain
inputs.  It is not clear how the precision/recall trade-off can be
controlled in the IBM models.

One advantage that Brown et al.'s Model 1 has over our word-to-word
model is that their objective function has no local maxima.  By using
the EM algorithm \cite{em}, they can guarantee convergence towards the
globally optimum parameter set.  In contrast, the dynamic nature of
the competitive linking algorithm changes the $\Pr(data | model)$ in a
non-monotonic fashion.  We have adopted the simple heuristic that the
model ``has converged'' when this probability stops increasing.

\section{Conclusion}
Many multilingual NLP applications need to translate words between
different languages, but cannot afford the computational expense of
modeling the full range of translation phenomena.  For these
applications, we have designed a fast algorithm for estimating
word-to-word models of translational equivalence.  The estimation
method uses a pair of hidden parameters to measure the model's
uncertainty, and avoids making decisions that it's not likely to
make correctly.  The hidden parameters can be conditioned on
information extrinsic to the model, providing an easy way to integrate
pre-existing knowledge.

So far we have only implemented a two-class model, to exploit the
differences in translation consistency between content words and
function words.  This relatively simple two-class model linked word
tokens in parallel texts as accurately as other translation models in
the literature, despite being trained on only one fifth as much data.
Unlike other translation models, the word-to-word model can
automatically produce dictionary-sized translation lexicons, and it
can do so with over 99\% accuracy.  

Even better accuracy can be achieved with a more fine-grained link
class structure.  Promising features for classification include part
of speech, frequency of co-occurrence, relative word position, and
translational entropy \cite{sement}.  Another interesting extension is
to broaden the definition of a ``word'' to include multi-word lexical
units \cite{smadja}.  If such units can be identified {\em a priori},
their translations can be estimated without modifying the word-to-word
model.  In this manner, the model can account for a wider range of
translation phenomena.

\section*{Acknowledgements}
The French/English software manuals were provided by Gary Adams of Sun
MicroSystems Laboratories.  The weather bitext was prepared at the
University of Montreal, under the direction of Richard Kittredge.
Thanks to Alexis Nasr for hand-evaluating the weather translation
lexicon.  Thanks also to Mike Collins, George Foster, Mitch
Marcus, Lyle Ungar, and three anonymous reviewers for helpful
comments.  This research was supported by an equipment grant from Sun
MicroSystems and by ARPA Contract \#N66001-94C-6043.

\end{document}